%% file: main_Arxiv.tex
\documentclass{article}
\PassOptionsToPackage{numbers}{natbib}
\usepackage[preprint]{neurips_2019}

\usepackage[utf8]{inputenc} 
\usepackage[T1]{fontenc}    
\usepackage{hyperref}       
\usepackage{url}            
\usepackage{booktabs}       
\usepackage{amsfonts}       
\usepackage{nicefrac}       
\usepackage{microtype}      
\usepackage{graphicx}
\usepackage{subcaption}
\usepackage{bm}
\usepackage{amsmath}
\usepackage{amssymb}
\usepackage{amsthm}
\usepackage{hyperref}
\usepackage{algorithm}
\usepackage{algorithmic}

\newcommand{\X}{\mathbf{x}}

\newcommand{\PHI}{\bm{\phi}}
\newcommand{\THETA}{\bm{\theta}}

\newtheorem{theorem}{Theorem}

\title{Deep Data Density Estimation through Donsker-Varadhan Representation}

\author{%
  Seonho Park $\:\:\:$ Panos M. Pardalos \\
  Department of Industrial and Systems Engineering \\ 
  University of Florida\\
  Gainesville, Florida, USA\\
  \texttt{\{seonhopark,pardalos\}@ufl.edu} \\
}

\begin{document}

\maketitle

\begin{abstract}
\input{sections/0abstract}
\end{abstract}

\section{Introduction}\label{sec:introduction}
\input{sections/1intro}

\section{Background} \label{sec:background}
\input{sections/2background}

\section{Methodology}\label{sec:methodology}
\input{sections/3methodology}


\section{Related Works}\label{sec:related_works}
\input{sections/4related_works}

\section{Experiments}\label{sec:experiments}
\input{sections/5experiments}

\section{Applications}\label{sec:applications}
\input{sections/6applications}

\section{Discussion}\label{sec:discussion}
\input{sections/7discussion}

\bibliography{ref}
\bibliographystyle{unsrtnat}


\end{document}

%% file: sections/0abstract.tex
Estimating the data density is one of the challenging problems in deep learning.
In this paper, we present a simple yet effective method for estimating the data density using a deep neural network and the Donsker-Varadhan variational lower bound on the KL divergence.
We show that the optimal critic function associated with the Donsker-Varadhan representation on the KL divergence between the data and the uniform distribution can estimate the data density. 
We also present the deep neural network-based modeling and its stochastic learning.
The experimental results and possible applications of the proposed method demonstrate that it is competitive with the previous methods and has a lot of possibilities in applied to various applications. 

%% file: sections/1intro.tex
Estimating the joint density $p(\X)$ of the data remains a particular challenge in unsupervised learning.
Strikingly, recent research on estimating the density using deep neural networks has made substantial progress on this research task.
There are two distinguishable ways to estimate the data density in deep learning: 1) Normalizing flows \cite{rippel2013high,papamakarios2017masked,papamakarios2019normalizing,dinh2014nice,dinh2016density,kingma2016improving} and 2) Autoregressive flows \cite{germain2015made,oord2016conditional,salimans2017,van2016pixel}.
Normalizing flow aims to model the invertible mapping from the data space to the basis distribution space and estimates the density using the changes of volume and basis density on the mapped point.
Autoregressive flow utilizes the conditional probability to model the function that is usually affected by the order of the conditioning. 
Both methods have theoretical foundations on the modeling the data density, and when training, minimizing negative log-likelihood is explicitly utilized.
The main concerns with these methods include how to assume the data distribution so as to make the model flexible enough to represent the complex and multi-modal data distribution elaborately.

In this paper, we suggest another avenue for estimating the data distribution by using the Donsker-Varadhan variational bound on the KL divergence, which is called \textbf{Deep Data Density Estimation (DDDE)} in the following.
The Donsker-Varadhan representation is a tight lower bound on the KL divergence, which has been usually used for estimating the mutual information \cite{hjelm2018learning,tschannen2019mutual,belghazi2018mutual} in deep learning.

We show that the Donsker-Varadhan representation of the KL divergence between the data and the uniform distribution can estimate the log probability of a data sample and has a sufficient ground on it in a theoretical way.
We also suggest the approach of training the model stochastically and show that estimating the density of the data through the DDDE is successful on the various datasets experimentally.
Finally, we show some applications of the DDDE including modifying the loss function in machine learning and anomaly detection.

\begin{figure*}[t!]
\centering
\vskip 0.2in
\begin{subfigure}[c]{0.24\textwidth}
\includegraphics[width=\textwidth]{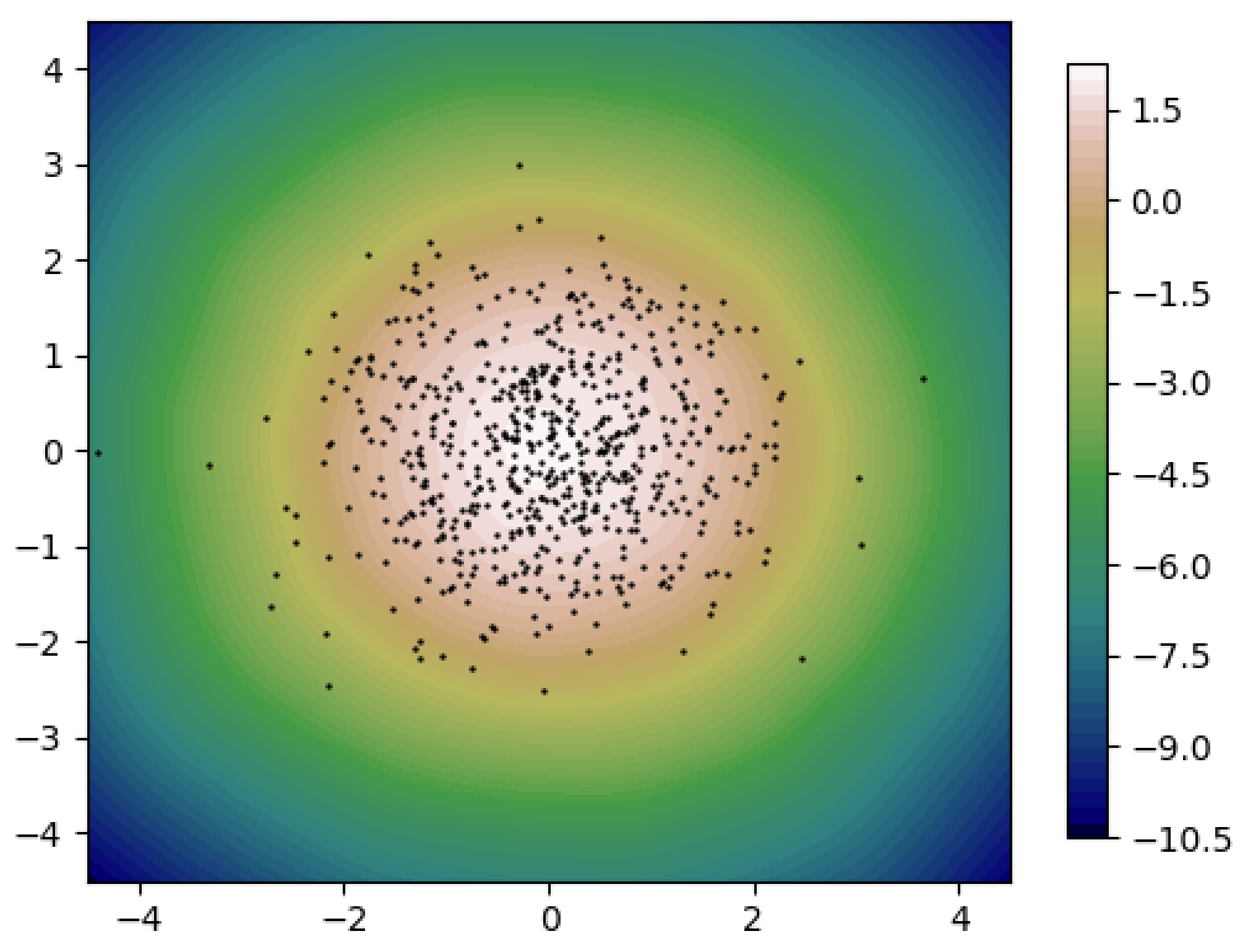}
\caption{}\label{subfig:gaussian0}
\end{subfigure}
\begin{subfigure}[c]{0.24\textwidth}
\includegraphics[width=\textwidth]{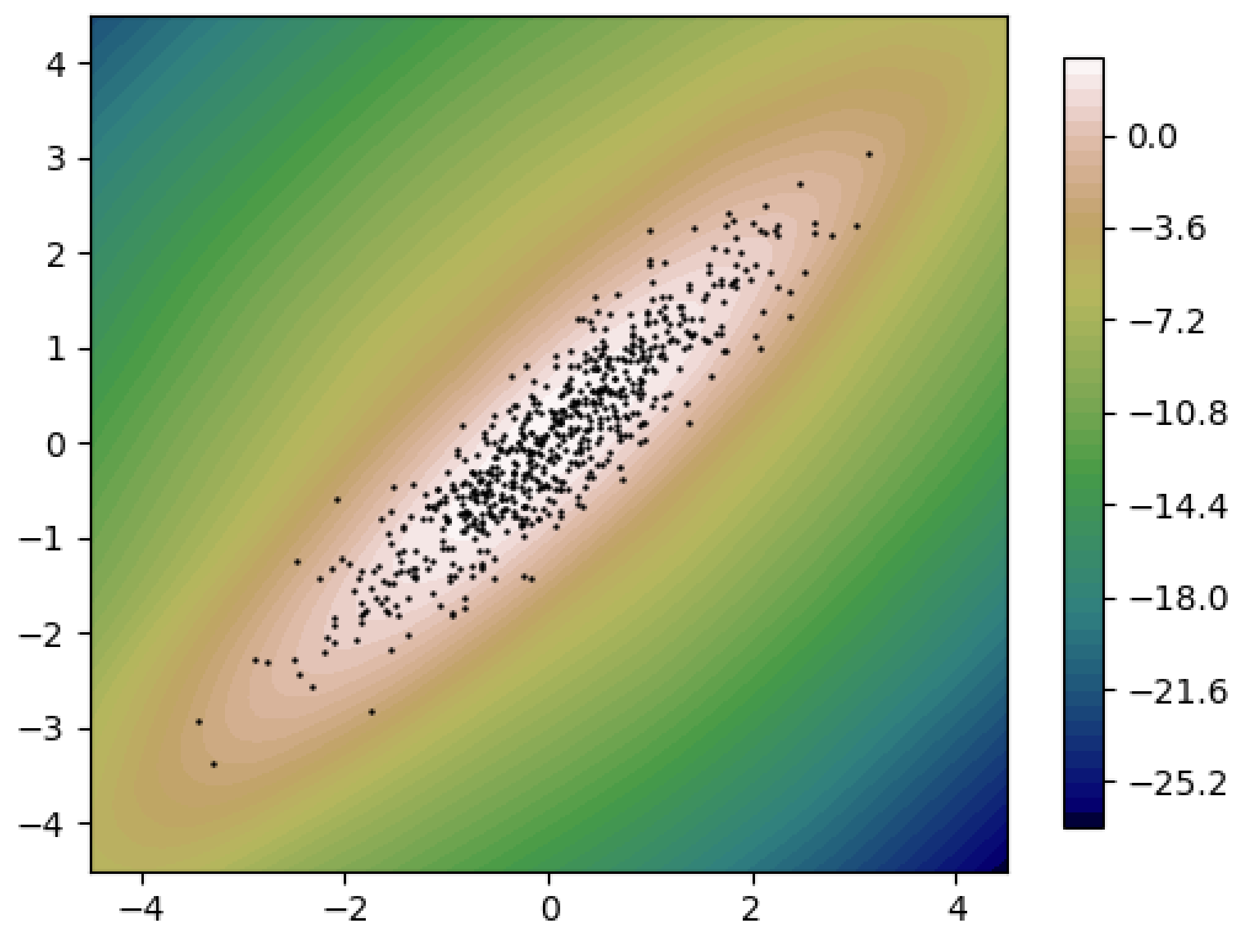}
\caption{}\label{subfig:gaussian9}
\end{subfigure}
\begin{subfigure}[c]{0.24\textwidth}
\includegraphics[width=\textwidth]{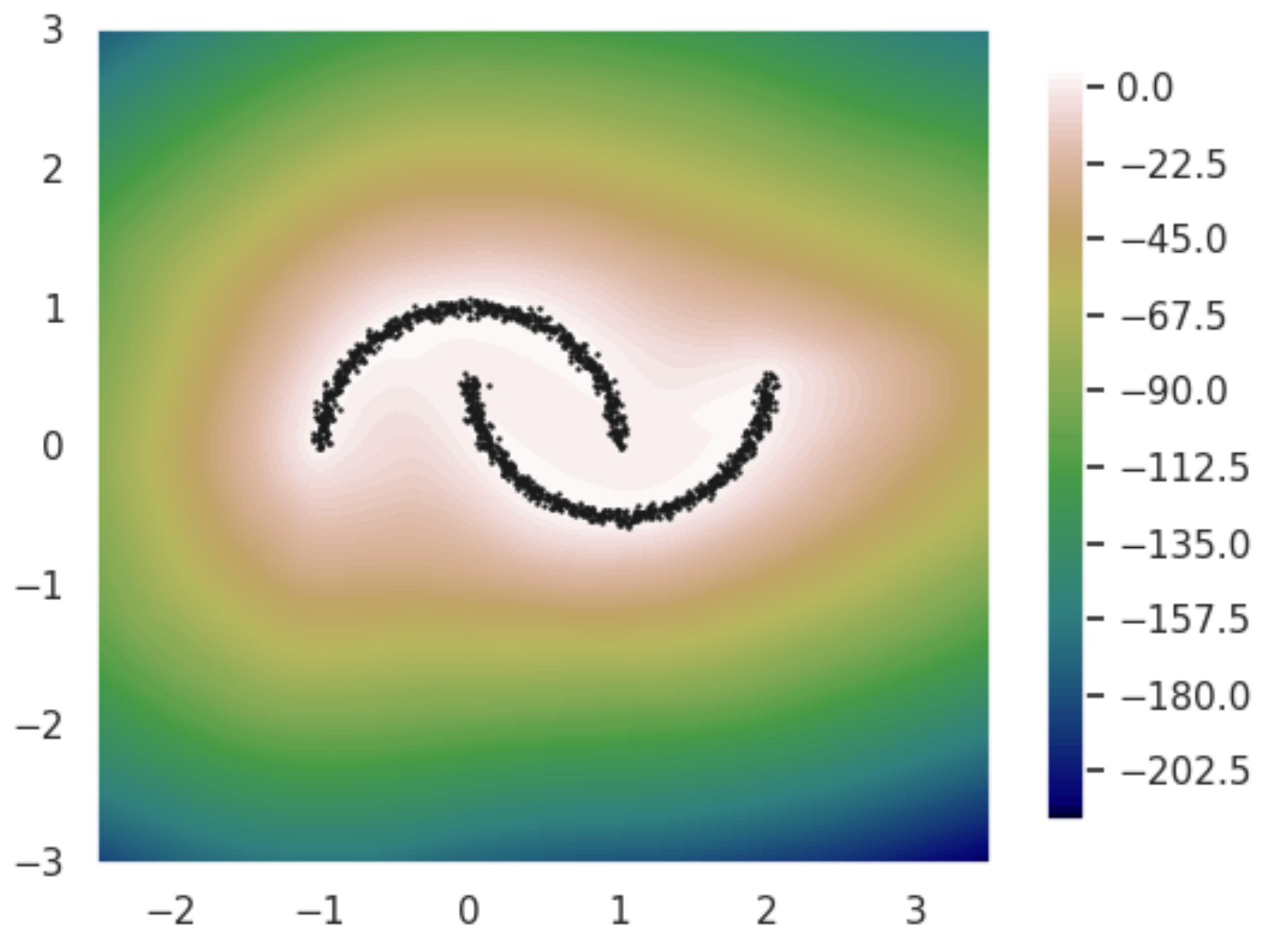}
\caption{}\label{subfig:twomoons}
\end{subfigure}
\begin{subfigure}[c]{0.24\textwidth}
\includegraphics[width=\textwidth]{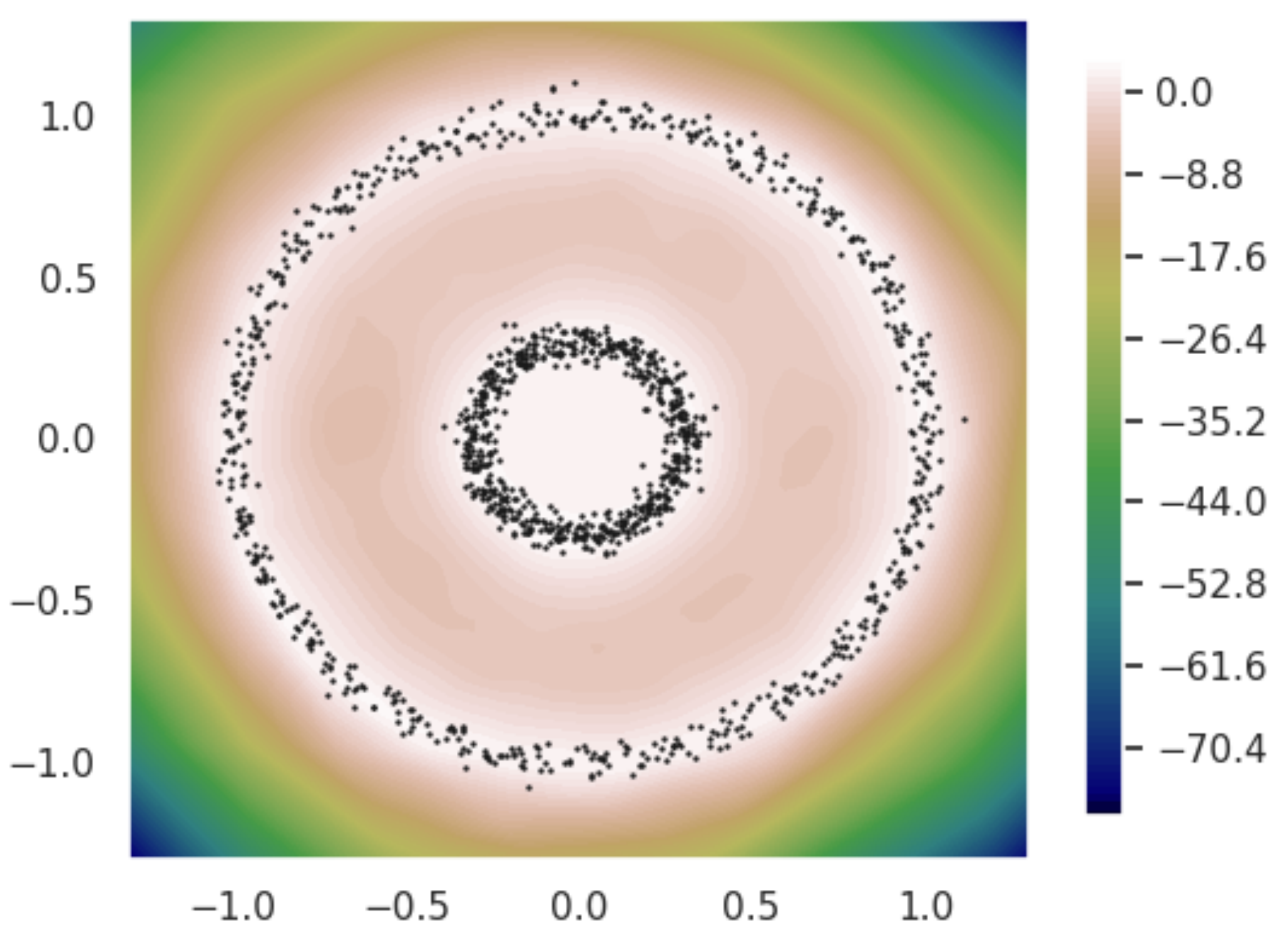}
\caption{}\label{subfig:circles}
\end{subfigure}
\vskip 0.2in
\caption{Contours represent the data density estimations by the proposed model, Deep Data Density Estimation (DDDE), on the scattered data points. (a) Isotropic Gaussian. (b) Correlated Gaussian with $\rho=0.9$. (c) Two moons data from \url{sklearn.datasets.make_moons}. (d) Circles data from \url{sklearn.datasets.make_circles}.
}\label{fig:ex_toy}
\vskip 0.2in
\end{figure*}

%% file: sections/2background.tex
\paragraph{KL Divergence}
In information theory, the Kullback-Leibler (KL-) divergence between two probability distributions $\mathbb{P}$ and $\mathbb{Q}$ of continuous random variables, which is defined on the same probability space $\mathcal{X}$, is
\begin{equation}\label{eq:dkl}
    D_{KL}(\mathbb{P}||\mathbb{Q}) := \mathbb{E}_\mathbb{P}\left[ \log \frac{p(\X)}{q(\X)} \right] = 
    \int_{\mathcal{X}} p(\X)\log\frac{p(\X)}{q(\X)}d\X,
\end{equation}
where $p(\X)$ and $q(\X)$ are the probability density functions of $\mathbb{P}$ and $\mathbb{Q}$, respectively, and $\frac{p(\X)}{q(\X)}$ is called the Radon–Nikodym derivative of $\mathbb{P}$ with respect to $\mathbb{Q}$.
$\mathbb{P}$ has to be absolutely continuous with respect to $\mathbb{Q}$ in order to define Eq.\ref{eq:dkl}.
The KL-divergence represents how $\mathbb{P}$ is different from $\mathbb{Q}$.
It is also known that the KL-divergence is always non-negative and $D_{KL}(\mathbb{P}||\mathbb{Q})=0$ if and only if $\mathbb{P}$ and $\mathbb{Q}$ are identical almost everywhere.

\paragraph{Donsker-Varadhan Representation}
Calculating the KL-divergence between the distributions that follow the well-known probability distributions such as Gaussian distribution is explicit and tractable to obtain.
However, when higher dimensional data is involved and the distribution underlying the data is unknown, it is quite difficult to measure the KL-divergence.
In order to resolve this issue, many researchers have researched the bounds of the KL-divergence \cite{belghazi2018mutual,poole2019variational,mcallester2020formal} that can be minimized or maximized given a family of the parameterized functions. 
Particularly, Donsker-Varadhan variational representation \cite{donsker1983asymptotic} that gives a tighter lower bound on the KL-divergence is used for this study.

\begin{theorem}[Donsker-Varadhan variational representation \cite{donsker1983asymptotic}]\label{thm:dv}
Given two probability measures $\mathbb{P}$ and $\mathbb{Q}$ over $\mathcal{X}$ with a finite KL-divergence. 
For all bounded functions $T(\X):\mathcal{X}\rightarrow \mathbb{R}$
\begin{equation}\label{eq:dv}
    D_{KL}(\mathbb{P}||\mathbb{Q}) = \sup_{T(\X):\mathcal{X}\rightarrow \mathbb{R}} \mathbb{E}_{\mathbb{P}}[T(\X)]-\log \mathbb{E}_{\mathbb{Q}}[e^{T(\X)}].
\end{equation}
Also the equality holds for some $T$.
\end{theorem}

Maximizing the right-hand side of Eq.\ref{eq:dv} gives an estimation of the KL divergence from $\mathbb{Q}$ to $\mathbb{P}$.
The issue associated in this problem is that we have to restrict the function family of $T$ in which we need to find the best function working in this equality.

Based on the lower bound of the KL-divergence shown in Eq.\ref{eq:dv}, the optimal function $T^*$ can be achieved by maximizing the Donsker-Varadhan lower bound, which is denoted as 
\begin{equation}\label{eq:opt_critic}
T^*(\X) = \log\frac{p(\X)}{q(\X)} + \log \mathbb{E}_{\mathbb{Q}}[e^{T^*}].    
\end{equation}
Deriving Eq.\ref{eq:opt_critic} is simple.
With a function $T$, suppose that there is a Gibbs distribution $\mathbb{G}$ defined on the same space of $\mathbb{Q}$ following $g(\X)=\frac{1}{\mathbb{E}_{\mathbb{Q}}[e^{T}]}e^Tq(\X)$ where $g(\X)$ is a probability density of $\mathbb{G}$.
Then taking logarithm and expectation with respect to $\mathbb{P}$ on both sides gives
\begin{align*}
    \mathbb{E}_{\mathbb{P}}\left[ \log\frac{g(\X)}{q(\X)} \right]  = \mathbb{E}_{\mathbb{P}}[T]-\log(\mathbb{E}_{\mathbb{Q}}[e^T]).
\end{align*}

Therefore, we can get
\begin{align*}
\mathbb{E}_{\mathbb{P}}\left[ \log\frac{g(\X)}{q(\X)} \right]  = -D_{KL}(\mathbb{P}||\mathbb{G}) + D_{KL}(\mathbb{P}||\mathbb{Q})\\
    D_{KL}(\mathbb{P}||\mathbb{G}) = D_{KL}(\mathbb{P}||\mathbb{Q})-(\mathbb{E}_{\mathbb{P}}[T]-\log(\mathbb{E}_{\mathbb{Q}}[e^T]))\geq 0
\end{align*}
The equality holds if and only if $\mathbb{P}=\mathbb{G}$ almost everywhere, so the optimal function $T^*$ has a form of Eq.\ref{eq:opt_critic}.

It is noted that in Eq.\ref{eq:opt_critic} the optimal function $T^*$ can be denoted as the logarithm of the Radon-Nikodym derivative of $\mathbb{P}$ with respect to $\mathbb{Q}$ plus the constant $\log \mathbb{E}_{\mathbb{Q}}[e^{T^*}]$, and also the Radon-Nikodym derivative represents the ratio of the probability densities between $p(\X)$ and $q(\X)$ given the measures. 

%% file: sections/3methodology.tex
\paragraph{Optimal Function $T^*$ of the Donsker-Varadhan Representation}
Now, we more focus on the optimal function $T^*$ of the Donsker-Varadhan variational bound.
Suppose that $p(\X)$ represents the data distribution that is basically unknown and it is likely defined on the high dimensional space.
In order to estimate $p(\X)$, we impose the uniform distribution $\mathcal{U}$ to $\mathbb{Q}$ over the same dimensional support as $\mathbb{P}$ and it is noted that it gives a constant probability density function. 
Then from Eq.\ref{eq:opt_critic}, the optimal function $T^*$ of the Donsker-Varadhan lower bound of the KL-divergence between $\mathbb{P}$ and $\mathcal{U}$ would be 
\begin{equation}\label{eq:optimal_T}
    T^*(\X) = \log p(\X) -\log u(\X)+ \log \mathbb{E}_{\mathcal{U}}[e^{T^*}] 
\end{equation}
where $u(\X)$ represents a constant probability density of the uniform distribution.
It is indicated that the optimal function $T^*$ can capture the log probability of $\mathbb{P}$ because the last two terms of the right hand side of Eq.\ref{eq:optimal_T} are constant.

\paragraph{Deep Neural Network Modeling}
When it comes to estimating the density of the higher dimensional data, deriving the optimal critic $T$ can be suffered from the curse of dimensionality.
Because the data lies on the very small region, saying data manifold, of the high dimensional data space and the probability density function of the uniform distribution $u(\X)$ is constant but minuscule over the support. 
Thus $T^*$ is to estimate notoriously small negative, i.e. $T^*\approx-\infty$, on the most of the high dimensional space, because $p(\X)\ll\epsilon$.
In order to mitigate this and also prevent the model from estimating too small negative on not so significant region of the support, the neural network model $f_{\THETA}$ aims to estimate $e^{T}>0$ instead of $T$ itself in the Donsker-Varadhan representation \ref{eq:dv}.
Thus we added the ELU activation \cite{clevert2015fast} with $\alpha=1$ at the output of the neural network model for our empirical experiments as well as applications shown in upcoming sections.
The ELU activation outputs the scalars greater than $-1$, thus we also added $1$ to enforce it to output the positive value, say, $f_{\THETA}(\mathbf{x}) = \text{ELU}(\mathbf{y})+1+\epsilon$, where $\mathbf{y}$ is the output of the neural network model and $\epsilon$ is the given small positive parameter that has a role to control the lower limit of the density value over the given data space.

Also, when it comes to stochastic estimation, the third term of the RHS in Eq.\ref{eq:optimal_T} is biased. 
To form a tractable unbiased estimation, based on the concavity nature of the logarithmic function, we used the following inequality: $\log(x) \leq \frac{x}{a}+\log(a)-1$ where $a>0$. This bound is tight when $x=a$. 
For the constant $a$ in our formulation, the exponential moving average of the expectation of $f_{\THETA}$ with respect to the uniform distribution, $\tilde{\mathbb{E}}_{\mathcal{U}}\left[f_{\THETA}\right]$, is used so that we expect to have a tight bound on the linearized form of $\log\mathbb{E}_{\mathcal{U}}[e^{T^*}]$.

Suppose that we have a dataset $\{\X_i\}_{i=1}^N$ that are i.i.d sampled, from the data distribution $\mathbb{P}$ defined over the finite and possibly high dimensional space.
Let us denote the estimated function as $f_{\THETA}(\mathbf{x})$, a neural network based function of $\mathbf{x}$, parameterized by $\THETA$.
Please note that this neural network based function $f_{\THETA}$ aims to estimate $e^T$ in the Donsker-Varadhan representation and it outputs a positive real scalar given each data point.
In order to estimate the data density function, we optimize $f_{\THETA}$ by maximizing the Donsker-Varadhan representation of the KL-divergence of $\mathbb{P}$ and $\mathcal{U}$ given the dataset as
\begin{equation}\label{eq:opt_dv}
    \max_{\THETA} \mathbb{E}_{\mathbf{x}\sim \mathbb{P}}[\log{f_{\THETA}(\mathbf{x})}]-\frac{\mathbb{E}_{\mathbf{x}'\sim\mathcal{U}}\left[ f_{\THETA}(\mathbf{x}') \right]}{\tilde{\mathbb{E}}_{\mathcal{U}}\left[f_{\THETA}\right]}-\tilde{\mathbb{E}}_{\mathcal{U}}\left[f_{\THETA}\right]+1
\end{equation}
Given the samples from the data $\X^{(i)}$ and the generated samples from the uniform distribution $\X'^{(i)}$, we can formulate a finite version of Eq.\ref{eq:opt_dv} as
\begin{equation}\label{eq:opt_dv2}
    \max_{\THETA} L_{\text{DV}}(\THETA) = \frac{1}{N_{\mathcal D}}\sum_{i}^{N_{\mathcal D}} \log f_{\THETA} (\mathbf{x}^{(i)}) - \frac{\left( \frac{1}{N_{\mathcal U}} \sum_i^{N_{\mathcal U}} f_{\THETA}(\mathbf{x}'^{(i)}) \right)}{\tilde{\mathbb{E}}_{\mathcal{U}}\left[f_{\THETA}\right]}-\tilde{\mathbb{E}}_{\mathcal{U}}\left[f_{\THETA}\right]+1,
\end{equation}
where $N_{\mathcal D}$ and $N_{\mathcal U}$ represent the number of samples from the data and the uniform distribution, respectively.
At every iteration, the exponential moving average $\tilde{\mathbb{E}}_{\mathcal{U}}\left[f_{\THETA}\right]$ is updated with the parameter $\beta$ as
\begin{equation}\label{eq:ma}
    \tilde{\mathbb{E}}_{\mathcal{U}}\left[f_{\THETA}\right] = \beta\tilde{\mathbb{E}}_{\mathcal{U}}\left[f_{\THETA}\right] + (1-\beta)\left( \frac{1}{N_{\mathcal U}} \sum_i^{N_{\mathcal U}} f_{\THETA}(\mathbf{x}'^{(i)}) \right).
\end{equation}
As a result, based on Eq.\ref{eq:optimal_T}, the optimal function $f_{\THETA}^*$ would approximate the true log probability as  
\begin{equation}
\log p(\X) \simeq \log f_{\THETA}^*(\X) + \log u(\X) - \tilde{\mathbb{E}}_{\mathcal{U}}\left[f_{\THETA}\right]
\end{equation} 
Therefore, we can say that the estimate function $f_{\THETA}(\mathbf{x})$ aims to estimate the log probability of the data, thus we call this function the DDDE in the following sections. 

The overall training process of the DDDE model is illustrated in Algorithm \ref{alg:algorithm}.
Opt$(\THETA, \eta, \Delta\THETA)$ in line 8 represents applying a stochastic optimization algorithm, where $\Delta\THETA$ represents the gradient of Eq.\ref{eq:opt_dv2} w.r.t. $\THETA$ or its variants.
Suppose that we handle a dataset containing a lot of samples then we need to sample a small subset of the dataset and apply stochastic gradient descent or its variant such as Adam \cite{kingma2014adam}, or a second-order method \cite{park2020combining} to train the DDDE model efficiently.

\begin{algorithm}[tb]
  \caption{Training DDDE}
  \label{alg:algorithm}
\begin{algorithmic}[1]
    \STATE Input: learning rate $\eta$, minibatch size $N_{\mathcal D}$, minibatch size $N_{\mathcal U}$ for the uniform distribution, positive offset $\epsilon$, exponential moving average parameter $\beta$.
    \STATE Calculate the constant log probability of the uniform distribution, $\log u(\X)$
    \REPEAT
    \STATE Draw $\mathbf{x}^{(1)},\dots,\mathbf{x}^{(N_{\mathcal D})}$ samples at random from the dataset
    \STATE Draw $\mathbf{x}'^{(1)},\dots,\mathbf{x}'^{(N_{\mathcal U})}$ samples from the uniform distribution
    \STATE Calculate the exponential moving average $\tilde{\mathbb{E}}_{\mathcal{U}}\left[f_{\THETA}\right]$ with Eq.\ref{eq:ma}.
    \STATE Model the loss function (Eq.\ref{eq:opt_dv2})
    \STATE Update the parameters $\THETA$ with \text{Opt}$(\THETA, \eta, \Delta\THETA)$
    \UNTIL{$\THETA$  has converged}
    \STATE Output: learned parameters $\THETA$ of the DDDE
\end{algorithmic}
\end{algorithm}

%% file: sections/4related_works.tex
\paragraph{Donsker-Varadhan Representation in Deep Learning}
One of the prevalent methods to utilize the Donsker-Varadhan variational representation in deep learning related research is Mutual Information Neural Estimation (MINE) \cite{belghazi2018mutual} that aims to estimate the mutual information between two distributions by using the Donsker-Varadhan representation. 
They focused on the fact that the mutual information can be represented as the KL-divergence between the joint distribution of two variables and the product of the marginal distributions. 
Then using stochastic estimation and neural network modeling arrive at MINE.
Another attempt to estimate the mutual information by means of the variational bounds including the Donsker-Varadhan representation is in \cite{poole2019variational,mcallester2020formal}.

Also, we would like to point out that $f$-divergence representation proposed in \cite{nguyen2010estimating}, which is a weaker bound than the Donsker-Varadhan representation can also be used for estimating the data density in the same way we have shown in the previous section.

\paragraph{Estimating Data Density}
One of the prevalent methods for estimating the data density through a nonparametric way is Kernel Density Estimation (KDE) \cite{parzen1962estimation}.
To estimate the density, KDE uses neighbor datapoints on the kernel space, parameterized by a bandwidth.
It often fails to estimate when the data dimensionality is high, because of its poor scalability and the curse of dimensionality \cite{ruff2018deep}.

Estimating the density of the high dimensional data is of a crucial problem in deep learning.
There are mainly two ways to estimate: 1) Autoregressive models, 2) Normalizing flows.
Autoregressive models such as MADE \cite{germain2015made}, PixelCNN \cite{oord2016conditional,salimans2017} and PixelRNN \cite{van2016pixel} estimate the probability density of the pixel given the known pixels by means of the conditional probability for images.
But our attempt is distinguished by the fact that DDDE directly estimates the probability density of the data itself whereas they were focused more on how to estimate the conditional probabaility of each feature of the data.
One of drawbacks of the autoregressive models is that it is sensitive to the order of the features for calculating the conditional probabilities.

Normalizing flow uses a bijective and invertible function mapping $f$ between the data space and the tractable base distribution.
The data density is estimated by calculating the density of a mapped point in the base distribution multiplied by the volume changes.
The main concern of the normalizing flow is how to design the tranformation function $f$, which is elaborated in \cite{papamakarios2019normalizing} that includes the previous attempts such as deep density model \cite{rippel2013high}, non-linear independent components estimation (NICE) \cite{dinh2014nice}, real non-volume preserving (Real NVP) \cite{dinh2016density}, inverse autoregressive flow (IAF) \cite{kingma2016improving}, masked autoregressive flow (MAF) \cite{papamakarios2017masked}.

Deep generative models also estimate the data density indirectly.
Variational autoencoder (VAE) \cite{kingma2013auto} aims to estimate the marginal distribution by the empirical lower bound (ELBO).
By doing so it creates the encoder-decoder structure to map from the data space to the latent space and \textit{vice versa}.
It is regarded as one of the prevalent deep generative methods to generate the data from the latent space, however, it can be used to estimate the probability density as shown in the previous works \cite{alemi2018fixing,park2021interpreting}, thus, it is also utilized to estimate the data density applications such as the anomaly detection \cite{an2015variational} as well.
Marginalizing the latent variable by means of importance sampling \cite{kingma2013auto} enables to estimate the density of the data, however, sampling is also involved in the computation so it is not straightforward.
It is also noted that the importance sampling of VAE is enhanced by attaching Inverse Autoregressive Flow (IAF) \cite{kingma2016improving} to the encoder of a VAE model.

%% file: sections/5experiments.tex
\paragraph{Estimating Density on Simple Distributions}
In order to verify the performance of the DDDE, we first test on the simple distributions such as two-dimensional Gaussian and Gaussian mixture distributions.
The support domain is restricted to $[0.,1.]^2$ and the uniform distribution for training the DDDE is defined on this domain space. 
The ground truth Gaussian distribution is isotropic and has a mean of $0.5$ and standard deviation of $0.1$.
The Gaussian mixture distribution has 9 isotropic Gaussian of which the center is located at grid points of $(0.2,0.5,0.8)$ and the standard deviation is $0.05$.
The neural network of the DDDE consists of 3-layered MLP with 512 hidden nodes.
The Adam optimizer \cite{kingma2014adam} is used with the learning rate of $0.001$, which is decreased by $10^{-\frac{1}{2}}$ at every 50 epochs and it is trained up to 200 epochs.
The number of total samples is $2048$ and at every iteration, $N_{\mathcal{D}}$ and $N_{\mathcal{U}}$ are $32$ and $64$, respectively.
The parameters $\beta$ and $\epsilon$ for the DDDE are set to $0.9999$ and $1\mathrm{e}-20$, respectively.
KDE is also tested as a benchmark for comparison. 
The bandwidth parameter $b$ of KDE are determined by 5-fold CV from $b\in\{2^{-5},2^{-4},\dots,2^0,\dots,2^4,2^5\}$.
The Gaussian type kernel is used because it is the best for estimating Gaussian or Gaussian mixture distributions in our experiments.

\begin{figure*}[ht]
\centering
\includegraphics[width=0.8\textwidth]{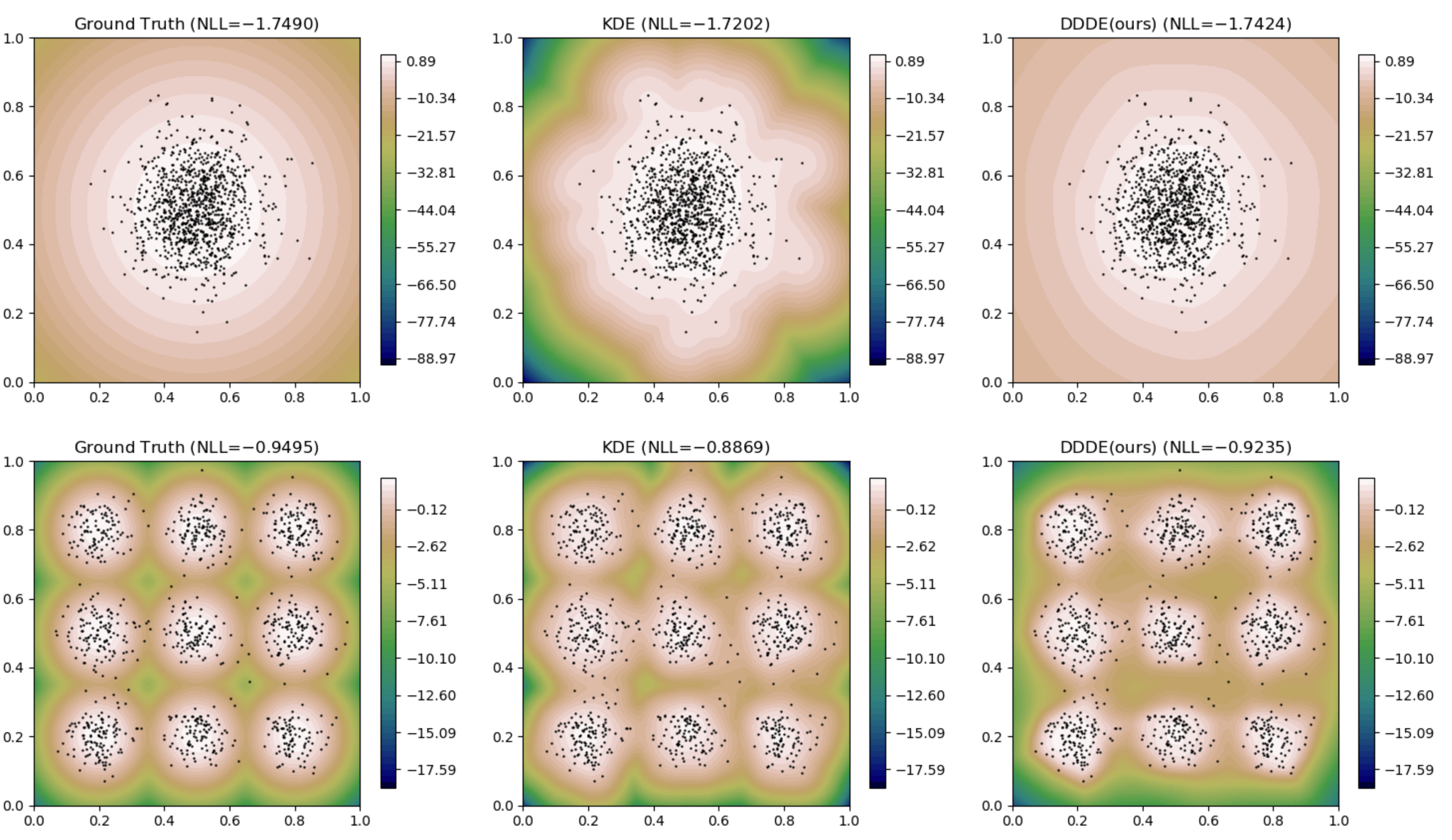}
\caption{Log-likelihood contours based on KDE and the DDDE and the negative log likelihood (NLL). The lower NLL is the better. \textbf{Top:} the Gaussian Distribution $\mathcal{N}(0.5,0.1)$. \textbf{Bottom:} the Gaussian mixture distribution.}\label{fig:NLL}
\vskip 0.2in
\end{figure*}

The result is measured by the negative log-likelihood (NLL) calculated on a testset of $10000$ samples.
As in Fig.\ref{fig:NLL}, the log-likelihood contours show that the DDDE estimates the density of the data well enough.
The NLL results show that the DDDE can estimate better than KDE in these cases. 

\paragraph{Estimating Density on Images}

\begin{figure*}[ht]
\centering
\includegraphics[width=0.9\textwidth]{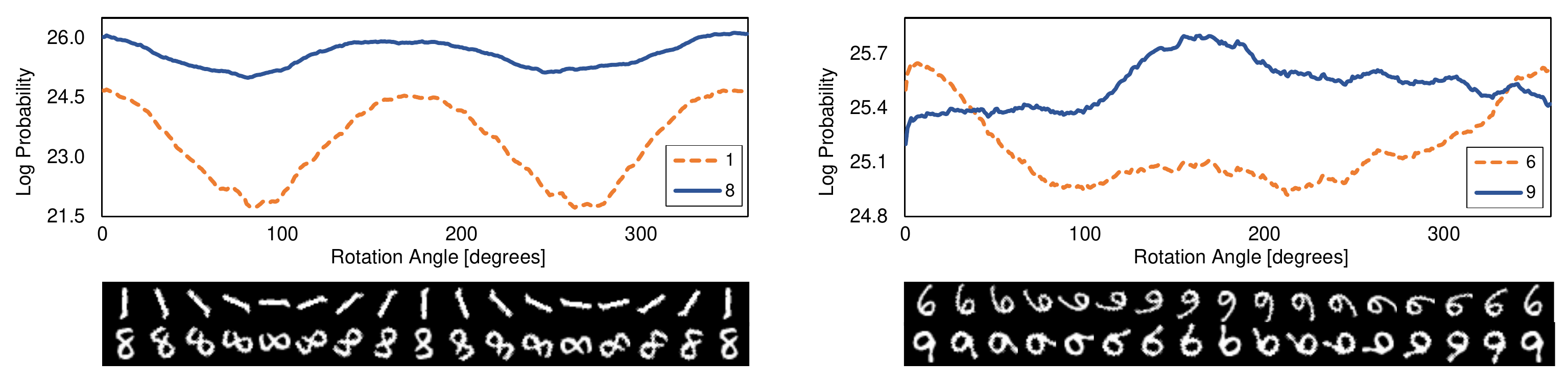}
\caption{\textbf{Left:} The DDDE is trained on 1's or 8's and tested on the rotated 1's and 8's, respectively. \textbf{Right:} The DDDE is trained on 6's and tested on the rotated 6's and 9's.}\label{fig:MNIST}
\vskip 0.2in
\end{figure*}

We trained the DDDE on the MNIST dataset \cite{lecun1998mnist} without any data augmentation except for normalizing to $[0.,1.]$.
As shown in Fig.\ref{fig:MNIST}, the DDDE was trained on a single type of digits in the MNIST dataset. 
For the left one in Fig.\ref{fig:MNIST} we trained two DDDE models on 1's and 8's digits, respectively, and tested on the rotated digits to look into the log probability estimations. 
The hyperparameter settings and the architecture of the DDDE are the same as the previous experiment.
The result shows that bottom-up digit of 1 and 8 looks same with the original digits, which can be detected by the log probability estimation of the DDDE. 
On the other hand, the right figure in Fig.\ref{fig:MNIST} shows the log probability results of the DDDE trained on 6's and the bottom-up 9 has a high probability because it looks like a digit 6.

%% file: sections/6applications.tex

%
\paragraph{Beyond the ERM Principle}
In supervised learning, suppose that the input data vector $\X$ and target $y$ follow the data distribution $P(X,Y)$.
We aim to find the function $f_{\PHI}\in\mathcal{F}$, parameterized by $\PHI$, to map from the input data vector $\X$ to the target $y$.
The difference between the function output and the target is penalized by the loss function $l$ over the data distribution.
The problem comes from the fact that the data distribution $P(X,Y)$ is intractable to calculate explicitly, thus we may rely on the realizations in the training dataset and use the \textit{empirical data distribution} $P_\delta(X,Y)$ to approximate $P(X,Y)$, which has a form of Dirac measures centered at the data points in the training dataset as $P_\delta(X,Y)$ = $\frac{1}{N}\delta(X=\X_i,Y=y_i)$ where $\{\X_i,y_i\}_{i=1}^N$ are datapoints in the training dataset. 
Thus the resultant risk function to be minimized through an optimization algorithm is defined as 
\begin{equation}\label{eq:erm}
\min_{\PHI} R(\PHI) = \frac{1}{N}\sum_{i=1}^{N} l(f_{\PHI}(\X_i),y_i).
\end{equation}
Learning the function $f_{\PHI}$ by minimizing Eq. \ref{eq:erm} is known as the empirical risk minimization (ERM) principle \cite{vapnik2013nature}, which is a canonical form for training machine learning models.

However, the ERM suffers from some issues including memorization \cite{simard1998transformation}, being vulnerable to the adversarial samples \cite{szegedy2013intriguing}. 
There are some previous research focused on data augmentation \cite{simard1998transformation}, interpolating augmenting samples \cite{zhang2017mixup}, employing variational information bottleneck \cite{belghazi2018mutual,alemi2016deep}, and vicinal risk minimization \cite{chapelle2001vicinal} to circumvent the issues of the ERM.  

Estimating the data distribution through the DDDE also can help improve the performance of learning the machine learning model and mitigating the generalization issues. 

Thus the risk function we wish to minimize is
\begin{equation}
    \min_{\THETA, \PHI} R(\THETA,\PHI) = \sum_{i=1}^N l(f_{\PHI}(\X_i), y_i)p_{\THETA}(\X_i,y_i)
\end{equation}
where $p_{\THETA}(\X_i,y_i)$ is a data density estimation through the DDDE.
We estimate the data distribution through the DDDE in two ways: unconditional and conditional.
The unconditional way is to estimate the data distribution $p(\X,y)$ as it is. 
Whereas the conditional way is to estimate the conditional distribution $p(\X|y)$ with the labels in the training dataset and at testing it estimates the joint distribution $p(\X,y)$ by multiplying the prior $\frac{1}{10}$ because the MNIST is to classify 10 handwritten digits.
For the classification model, we follow the same protocols including the training/test data splitting, data preprocessing, optimization and model configuration in \cite{alemi2016deep}.
For the DDDE model, we use the same architectures as well as the hyperparameters in Sec. \ref{sec:experiments}.

\begin{table}[t!]
\begin{center}
\begin{tabular}{@{}l|c@{}}
\toprule
Model    & Error rate {[}\%{]} \\ \midrule
Baseline (ERM) & 1.38\%\\ 
DropOut (\cite{pereyra2017regularizing})  & 1.40\%\\ 
Label smoothing (\cite{pereyra2017regularizing}) & 1.23\% \\
Confidence penalty (\cite{pereyra2017regularizing}) & 1.17\% \\
VIB (\cite{alemi2016deep})& 1.13\% \\
MINE (\cite{belghazi2018mutual})& 1.01\% \\\midrule
DDDE (unconditional) & 1.07\% \\ 
DDDE (conditional) & 0.98\% \\ \bottomrule
\end{tabular}
\end{center}
\caption{Permutation invariant MNIST misclassification rate. The protocol is from the previous works \cite{alemi2016deep,belghazi2018mutual,pereyra2017regularizing} for comparison purpose. The misclassification rates in the top columns are excerpted from the previous literature \cite{belghazi2018mutual} and the DDDE results in the bottom columns are mean values of ten independent runs with different random seeds. }\label{table:mnist}
\end{table}

The results in Table \ref{table:mnist} show the misclassification rates of the DDDEs along with the regularization methods.
The results support that the DDDE estimates the data density of the samples so that it imposes higher importance weights on the samples of which the probability lying on the data distribution is high, and as a consequence, it prevents the classification model from learning on the adversarial samples that may be contained in the training dataset as well. 

\paragraph{Anomaly Detection}

\begin{table}[t!]
\begin{center}
\begin{tabular}{@{}l|ccccccc@{}}
\toprule
Dataset & GPND & DSVDD & AE & VAE & AAE & PGN & DDDE (ours) \\ \midrule
MNIST   & 75.0\% & 93.5\% & 91.2\% & 90.1\% & 89.9\% & 94.7\% & 88.6\% \\ 
FMNIST  & 83.2\% & 91.4\% & 87.7\% & 86.9\% & 87.3\% & 91.5\% & 87.1\%\\ 
CIFAR-10 & 59.4\% & 61.7\% & 57.0\% & 58.2\% & 57.1\% & 64.4\% & 58.5\%\\ \bottomrule
\end{tabular}
\end{center}
\caption{Average AUROC performance. The results for GPND, DSVDD, AE, VAE, AAE, PGN are from \cite{park2021interpreting}. }\label{table:anomaly_detection}
\end{table}

Anomaly detection is the problem of identifying outliers that are usually unseen when training. 
Here, we used the same experiment protocol in \cite{park2021interpreting} on the MNIST, Fashion MNIST (FMNIST) \cite{xiao2017fashion}, and CIFAR-10 \cite{krizhevsky2014cifar} datasets.
The DDDE model contains three convolutional layers, which have the same configurations in the previous work, followed by the two fully connected layers to produce a scalar output. 
The dropout layer is attached at the penultimate layer with the dropout probability of $0.5$.
The minibatch sizes, $N_{\mathcal{D}}$ and $N_{\mathcal{U}}$ are $200$ and $500$, respectively.
Other experiment settings follow the previous research for a fair comparison.
The reciprocal of the log probability estimation through the DDDE is used as an anomaly score indicating that an instance with a higher score is prone to be anomalous. 
The result in Table \ref{table:anomaly_detection} shows that the density estimation through the DDDE can be competitively used as a measure of anomaly detection, although it is slightly worse than the state-of-the-art methods.

%% file: sections/7discussion.tex
In this work, we propose a new way of estimating the data density using deep neural networks, the DDDE that utilizes the Donsker-Varadhan variational bound on KL divergence.
The DDDE uses samples from the dataset and the uniform distribution that shares the same domain space of the data to estimate the data density.
This method sheds light on the way of estimating data density using deep neural networks that is basically different from previous popular approaches: normalizing flow and autoregressive flow.

Although it gives us the high possibility to estimate the data density, it also contains some drawbacks which can derive some research directions.
For example, when it comes to drawing the samples from the uniform distribution, it can also suffer from the curse of dimensionality because it has to sufficiently cover the region over the domain space evenly. 
Some possible ways to resolve include drawing the sample from a smaller space and map evenly to the higher space by means of the deep generative model or generating samples at the boundary of the data distribution as in \cite{lee2017training}, however, it needs some theoretical foundations.
The performance of the DDDE in this paper is not ultimately optimized, which means that adopting the higher level of the architecture, and scrutinizing the hyperparameters selection can help enhance the performance of the DDDE potentially.